\documentclass[letterpaper, 10 pt, conference]{ieeeconf}  
\pdfoutput=1

\IEEEoverridecommandlockouts                              

\usepackage{graphics} 
\usepackage{epsfig} 
\usepackage{mathptmx} 
\usepackage{times} 
\usepackage{amsmath} 
\usepackage{amssymb}  
\usepackage{multirow}
\usepackage{soul,color}
\usepackage{hyperref}
\soulregister\cite7
\soulregister\ref7
\soulregister\pageref7
\title{\LARGE \bf
Excitation Trajectory Optimization for Dynamic Parameter Identification Using Virtual Constraints in Hands-on Robotic System
}

\author{Huanyu Tian$^{1,2}$,  Martin Huber$^{1}$, Christopher E. Mower$^{1}$, Zhe Han$^{1,2}$, \\Changsheng Li$^{2}$, Xingguang Duan$^{2}$ and Christos Bergeles$^{1}$
\thanks{This work has been submitted to the IEEE for possible publication. Copyright may be transferred without notice, after which this version may no longer be accessible. 
This work was supported by Beijing Institute of Technology under Chasing Dream Abroad Scholarship, and from the EU's Horizon 2020 research and innovation program [agreement No 101016985 (FAROS project)].}
\thanks{$^{1}$ H.~Tian, M.~Huber, C.~E.~Mower, Z.~Han, and C.~Bergeles are with the School of Biomedical Engineering \& Imaging Sciences, King’s College London, UK. 
        {\tt\small christos.bergeles@kcl.ac.uk}}%
\thanks{$^{2}$ H.~Tian, Z.~Han, C.~Li, and X.~Duan are with the School of Mechatronical Engineering, Beijing Institute of Technology, Beijing 100081, China, and with the Key Laboratory of Biomimetic Robots and Systems, Beijing Institute of Technology, Ministry of Education, China}%
}

\begin{document}

\maketitle
\thispagestyle{empty}
\pagestyle{empty}

\begin{abstract}

This paper proposes a novel, more computationally efficient method for optimizing robot excitation trajectories for dynamic parameter identification, emphasizing self-collision avoidance. 
This addresses the system identification challenges for getting high-quality training data associated with co-manipulated robotic arms that can be equipped with a variety of tools, a common scenario in industrial but also clinical and research contexts. 
Utilizing the Unified Robotics Description Format (URDF) to implement a symbolic Python implementation of the Recursive Newton-Euler Algorithm (RNEA), the approach aids in dynamically estimating parameters such as inertia using regression analyses on data from real robots. 
The excitation trajectory was evaluated and achieved on par criteria when compared to state-of-the-art reported results which didn't consider self-collision and tool calibrations. Furthermore, physical Human-Robot Interaction (pHRI) admittance control experiments were conducted in a surgical context to evaluate the derived inverse dynamics model showing a 30.1\% workload reduction by the NASA TLX questionnaire.
\end{abstract}

\section{Introduction}
\label{Section1}
Autonomy and shared autonomy in human/robot environments is attracting significant research attention, both in industrial \cite{SelvaggioSharedControl,Mower2022An} and clinical/surgical settings \cite{AttanasioAutonomyInRobotics,yang2017medical}. Intuitive co-manipulation of a robotic arm by a human is a crucial step towards collaborative human/robot interactions. Effortless co-manipulation of robotic arms with high inertia and mass relies on precise identification of the robot's dynamic parameters. In addition, achieving an effective system identification solution for robotic systems with varying end effectors (EE) necessitates the selection of an appropriate criterion and the incorporation of self-collision constraints.

Dynamic parameter identification is a key pillar of research in pHRI, spanning both industrial\cite{wang2022design} and medical domains\cite{HongbinGravityCompensation}. This process is capable of transforming robotic dynamics, as represented by the Recursive Newton Euler Algorithm (RNEA), into a linear system of dynamic parameters \cite{sousa2019inertia}. The linearly represented dynamics can be simplified into an equation that only contains the dynamic parameters after substituting measured positions and forces into the dynamics over an excitation trajectory. Therefore, this leads to a conventional least squares regression problem through which the dynamic parameters can be identified.

In identification, the recorded states (positions, velocities, and accelerations) affect the robustness of parameter identification \cite{su2020deep}. To improve the identification quality, robots should execute a well-designed trajectory to excite the design matrix of the least squares regression and make the matrix perform well in the Fisher-information sense. Trajectory generation actuates every joint according to a pre-calculated joint-space plan to record joint positions, velocities, accelerations, and raw torques. By constructing the regression loss and considering the physical feasibility of parameters, semideﬁnite programming (SDP) can be applied to find optimal parameters\cite{wen2022dynamic}.



\begin{figure}[t]
	\centering
	\includegraphics[width=\columnwidth]{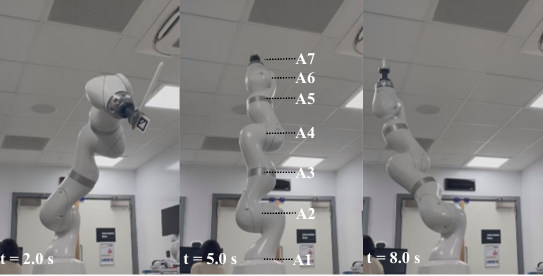}
     \vspace{-10pt}
	  \caption{Snapshots of the trajectory execution for the KUKA LBR Med7 R800 with an EE for different time steps. A1-A7 highlight the individual joints. Our method incorporates the EE's geometry to find trajectories that are collision-free and optimal for dynamic parameter identification.}
	  \label{figure1}
    \vspace{-10pt}
\end{figure}
We use the KUKA LBR Med7 R800 as our testbed and example in the narrative. The 7 DoF robot is equipped with torque sensors integrated into each joint and adheres to ISO standards related to incorporation into medical devices. It has been equipped with a variety of EEs, for laser bone cutting \cite{honigmann2022cold}, bone drilling \cite{siddiqi2022diagnosis}, dexterous endoscopic surgery \cite{virtuosorobotics:roboarmpro}, flexible endoscopy \cite{atlasendoscopy} and more. 
As a result, there is considerable research and development conducted on the KUKA LBR Med7 R800, with co-manipulation being a key focus. In \cite{sturz2017parameter}, Yvonne et al. introduced constraints for parameter regression to obtain feasible parameter sets, leading to a remarkably accurate estimation of the robot's dynamic model. However, their work was limited to an individual EE. To identify a system with arbitrary EEs, it is necessary to consider self-collision avoidance \cite{zhang2021obstacle} as collision-free excitation trajectories have to be guaranteed. 


Optimal excitation trajectories play a crucial role in achieving accurate system identification. The concept of optimality was extensively explored in \cite{lee2021optimal}, which revealed the connection between the Fisher information matrix and trajectory optimization. In essence, the Fisher information matrix quantifies the sensitivity of the dynamics of the model to changes in the parameter being estimated. This is often assessed by examining the eigenvalues of the inverse estimator covariance matrix. To delve deeper, several widely recognized criteria are commonly used: A-optimality, D-optimality, E-optimality, and the condition number \cite{braatz1994minimizing}.

Among them, the condition number offers the highest versatility in describing the quality of regression results. Yet, computing the condition number poses challenges due to its non-convex and periodic nature. While numeric Singular Value Decomposition (SVD) \cite{ayusawa2017generating, xu2020dynamic} or heuristics \cite{tika2020dynamic} can be applied, obtaining gradients becomes difficult, particularly when constraints are included in the optimization. 

This paper presents a pipeline for increased trajectory-execution safety and computational efficiency. We contribute:
\begin{itemize}
\item A novel approach for system identification based on data collected by performing optimized excitation trajectories incorporating collision avoidance. 
\item A cost function tailored for dynamic parameter regression, derived for the optimal trajectory design.
\item An open-source 
Python implementation of our approach that, given a URDF, can be applied to other robot arms.
\end{itemize}

Hardware realization of our method including a pHRI study, quantitative comparisons, and several evaluations demonstrate its performance. 




\section{Dynamic Model and Linearization Method}
\label{Section2}
This section describes the linear representation of robots' dynamics and the minimal inertia set decomposition.

\subsection{Model structure and compensation method}

To facilitate smooth and intuitive physical pHRI, we must account for Coriolis force, friction force, and gravity force to enhance force perception by the robot. The compensation of these forces can be seamlessly integrated with a feedforward model, which is the model that we seek to estimate. During this compensation, the external torque $\mathbf{\tau}_{ext}$ can be estimated using (\ref{equ1}):
\begin{equation}
\mathbf{\tau}_{ext} = \mathbf{\tau}_{raw} - \mathbf{\tau}_{m}
\label{equ1}
\end{equation}

Here, $\mathbf{\tau}_{raw}$ is provided by the robot's internal force sensor, while $\mathbf{\tau}_{m}$ represents the compensation model.
With the RNEA method, the dynamics of a manipulator can be described by (\ref{equ2}):
\begin{equation}
\mathbf{\tau}_{m} = \mathbf{M}(\mathbf{q})\mathbf{\ddot q} + \mathbf{C}(\mathbf{q, \dot q})\mathbf{\dot q} +\mathbf{g}(\mathbf{q}) + \mathbf{f_{st}}sgn(\mathbf{\dot q})+\mathbf{f_{v}}\mathbf{\dot q}
\label{equ2}
\end{equation}

Here, the dynamic model is defined with respect to the robot inertia matrix $\mathbf{M}(\mathbf{q})$, Coriolis matrix $\mathbf{C}(\mathbf{q, \dot q})$, and gravity vector $\mathbf{g}(\mathbf{q})$. Vector $\mathbf{q}$ represents the joint positions, while vector $\mathbf{\dot q}$ represents joint velocities. Joint acceleration is denoted by $\mathbf{\ddot q}$. Symbols $\mathbf{f_{st}}$ and $\mathbf{f_{v}}$ denote the Coulomb Friction coefficient, and Viscous Friction coefficient, respectively, while $sgn(\cdot)$ is the sign function.

\subsection{Dynamics linear representation and minimal inertia set}
From the representation shown in (\ref{equ2}), the parameter variables are expressed implicitly in dynamics. To address this, it's common to extract all parameters to be identified and reform the dynamics to (\ref{equ3}):
\begin{equation}
\mathbf{\tau}_{m} = \mathbf{\bar Y}(\mathbf{q},\mathbf{\dot q},\mathbf{\ddot q}) \theta
\label{equ3}
\end{equation}
where $\theta$ and $\mathbf{\bar Y}(\mathbf{q},\mathbf{\dot q},\mathbf{\ddot q})$ can be derived by the mentioned parameters extraction, which substitutes constant values like 1 or 0 with the motion states of dynamics (or its derivation) to find the coefficients of every dynamic parameter.
Notably, vector $\theta$ refers to a constant vector, which reflects the inherent properties of the robotic arm and its mounted EE, such as inertia, masses, and frictions of every link. 
Vector $\theta$ can be expressed as $\theta = \theta(\mathbf{M}(\cdot),\mathbf{C}(\cdot),\mathbf{g}(\cdot))$, which depends on the motion states represented by $\mathbf{q}$, $\mathbf{\dot q}$, and $\mathbf{\ddot q}$.
However, identification for $\theta$ leveraging least squares regression may be impossible as $\mathbf{\bar Y}(\mathbf{q},\mathbf{\dot q},\mathbf{\ddot q})$ is rank-deficient and has singularities. Therefore, it is necessary to reform $\theta$ into minimal inertia sets, (dependent parameters) $\theta_b$, which makes the regression matrix full-rank. First, we define the constant factor matrix $\mathbf{P}_b$ via (\ref{equ4b}):
\begin{equation}
\mathbf{\bar Y}_b = \mathbf{\bar Y} \mathbf{P}_b
\label{equ4b}
\end{equation}

After multiplying the constant factor matrix, matrix $\mathbf{\bar Y}_b$ can be full-rank, which is beneficial for regression. QR decomposition is applied to $\mathbf{\bar Y}$, thereby grouping together the vector bases corresponding to the decomposed principal elements as $\mathbf{P}_b$, while keeping the remaining vector bases separate as $\mathbf{P}_d$.
After obtaining constant factor matrices $\mathbf{P}_b$ and $\mathbf{P}_d$, the $\tau_m$ can be represented as follows:
\begin{equation}
\begin{aligned}
\tau_m &= \mathbf{\bar Y} \theta
=\mathbf{\bar Y}[\mathbf{P}_b, \mathbf{P}_d][\mathbf{P}_{b}^{T}, \mathbf{P}_{d}^{T}]^{T}\theta\\
&=\mathbf{\bar Y} \mathbf{P}_b (\mathbf{P}_b^{T} + \mathbf{P}_{b}^{-1}\mathbf{P}_{d} \mathbf{P}_d^{T})\theta=\mathbf{\bar Y}\mathbf{P}_{d}\mathbf{K} \theta
\end{aligned}
\label{equ5}
\end{equation}
In this context, the matrix $\mathbf{P}_b$ serves as a selector, identifying linearly independent basis elements within $\mathbf{\bar Y}$, while $\mathbf{P}_d$ signifies the presence of redundant basis components. Thus, the transformation of the basis from $\mathbf{\bar Y}$ to $\mathbf{\bar Y_b} = \mathbf{\bar Y} \mathbf{P}_b$ is accomplished. To facilitate concise notation, $\mathbf{K} = \mathbf{P}_b^{T} + \mathbf{K}_d \mathbf{P}_d^{T}$ is defined, where $\mathbf{K}_d = \mathbf{P}_{b}^{-1}\mathbf{P}_{d}$ is given.
Finally, the dependent inertia set is given $\theta_b = \mathbf{K}\theta$.



\section{Optimal Trajectory Generation}
\label{Section3}
As Sec.~\ref{Section1} mentioned, 
to reduce the sensitivity of identification, the matrix $\mathbf{\bar Y_b}$ (or $\mathbf{\bar Y_b}^{T} \mathbf{\bar Y_b}$ as matrix $\mathbf{\bar Y_b}$ is a non-square matrix) should be well-designed and the coefficients of the target trajectory 
are optimized accordingly.

\subsection{Optimal Excitation Trajectories for manipulators}
When estimating dynamic parameters, regression accuracy relies on the condition number of $\mathbf{\bar Y}_b$ represented by the ratio of its largest to smallest singular value.

However, obtaining a closed-form solution for the gradient of $cond(\mathbf{\bar Y_b})$ with respect to joint positions in symbolic libraries like CasADi \cite{andersson2019casadi} and OpTaS \cite{mower2023optas} is not computationally efficient. In addition, the numerical gradient solution of condition numbers, which relies on Singular Value Decomposition (SVD), involves numerical estimation that is susceptible to numerical instability. 


To address this problem, another criterion using closed-form gradients can be given. First, we can use the Gramian matrix $\mathbf{\bar H_b}(\mathbf{q},\mathbf{\dot q},\mathbf{\ddot q}) = \mathbf{\bar Y_b}^T(\mathbf{q},\mathbf{\dot q},\mathbf{\ddot q})\mathbf{\bar Y_b}(\mathbf{q},\mathbf{\dot q},\mathbf{\ddot q})$ as a substitute for the original matrix $\mathbf{\bar Y}_b$ in calculations. 
The Gramian matrix is a positive semidefinite matrix, and its eigenvalues are the squares of the eigenvalues of the original matrix.


\textbf{Theorem 1}: Given the trajectory optimization requirements and manipulator dynamics, consider the Frobenius norm sum of the Gramian matrix $\mathbf{\bar H_b}$ and its inverse matrix $\mathbf{\bar H_b^{-1}}$, denoted as $r_c = \frac{1}{2}(\mathbf{ ||\bar H_b||_{fro} + ||\bar H_b^{-1}||_{fro}})$. It can be established that $r_c$ serves as an upper bound for the condition number of the original matrix $\mathbf{\bar Y}_b$.
\begin{equation}
\begin{aligned}
r_c(\mathbf{q},\mathbf{\dot q},\mathbf{\ddot q})
= \mathop{sup}\;\; cond(\mathbf{\bar Y_b (\mathbf{q},\mathbf{\dot q},\mathbf{\ddot q})})
\label{equ6}
\end{aligned}
\end{equation}

\textbf{Proof}:
Let $\lambda_{max}$ and $\lambda_{min}$ be the largest and smallest eigenvalues of $\mathbf{\bar Y_b}$. By the properties of the Gramian matrix, we have that the largest and smallest eigenvalues of the Gramian matrix are $\lambda_{max}^{2}$, and $\lambda_{min}^{2}$, respectively. By the definition of Frobenius norm, we have:
\begin{equation}
\begin{aligned}
||\bar H_b||_{fro} = \sqrt{\sum_{i=1}^{m} \sum_{j=1}^{n} h_{ij}^2} = \lambda_{max}^{2}
\label{equ7}
\end{aligned}
\end{equation}
where $h_{ij}$ is an arbitrary element of $\bar H_b$. Similarly, for the inverse Gramian matrix $\bar H_b^{-1}$, we have $||\bar H_b^{-1}||_{fro} = 1/\lambda_{min}^{2}$.

Therefore, the cost function $r_c$ can be written as:
\begin{equation}
\begin{aligned}
r_c = \frac{1}{2}(\lambda_{max}^{2} + 1/\lambda_{min}^{2}) \geq \frac{\lambda_{max}} {\lambda_{min}} = cond(\mathbf{\bar Y_b (\mathbf{q},\mathbf{\dot q},\mathbf{\ddot q})})
\label{equ8}
\end{aligned}
\end{equation}
As a result, $r_c$ is the upper bound of the original matrix's condition number.
\hfill$\square$

We can use the upper bound of the condition number to obtain a solution to the optimization problem in a computationally efficient manner. Assuming $n$ denotes recorded positions here, $r_c$ has less computational complexity, i.e., $\mathcal{O}(n)$, compared to $\mathcal{O}(n^2)$ of $cond = \sqrt{||\bar H_b||_{fro} ||\bar H_b^{-1}||_{fro}}$. Thus, $r_c$ can be  a computationally efficient objective function; as $r_c$ decreases, the condition number will decrease accordingly.

The target trajectory can be represented with a $L$th order Fourier series with respect to cosine coefficients $a_{i,j}$ and sine coefficients $b_{i,j}$ shown in (\ref{equ9b}), where $i$ and $j$ denotes joint index and term number in Fourier series. This parameterization ensures cycle consistency, i.e., that a trajectory starts and finishes at the same point. The aim of our optimization is to find the optimal solution of $a_{i,j}$ and $b_{i,j}$.

As a result, we have the optimization problem:
\begin{subequations}
\begin{align}
\mathop{min}_{a_{i,l}, b_{i,l}}\;& r_c (\mathbf{q},\mathbf{\dot q},\mathbf{\ddot q}),  \label{equ9a} \\
\text{s.t.} \;&q_{i}:=\sum_{l=1}^{L}\frac{a_{i,l}}{\omega_{f}l} sin(\omega_{f}lt)-\frac{b_{i,l}}{\omega_{f}l}cos(\omega_{f}lt), \label{equ9b}\\ 
&[\sum_{l=1}^L \frac{a_{i,l}}{l}, \sum_{l=1}^L b_{i,l}, \sum_{l=1}^L a_{i,l}l]^{T} = 0, \label{equ9c}\\ 
&\sum_{l=1}^{L} \frac{1}{l} \sqrt{a_{i,l}^2 +b_{i,l}^2} \leq \omega_{f} q_{i,max}, \label{equ9d} \\
&\sum_{l=1}^{L} \sqrt{a_{i,l}^2 +b_{i,l}^2} \leq \dot q_{i,max}, \label{equ9e}\\
&[a_{i,l} b_{i,l}] \leq \mathop{min}(\frac{1}{L}\omega_{f}lq_{i,max}, \dot q_{i,max}) [1 1]^{T}, \label{equ9f}\\
&[a_{i,l} b_{i,l}] \geq \mathop{max}(\frac{1}{L}\omega_{f}lq_{i,min}, \dot q_{i,min}) [1 1]^{T}, \label{equ9g}
\end{align}
\end{subequations}
$\forall i = 1, ... 7$ (joints), $\forall l = 1,...5$ (order of terms in series), where the target trajectory \{$\mathbf{q}$ \vline $q_i, \;\;\forall i = 1, ... 7$\} is parameterized with a Fourier series. The order of the Fourier series is set to $L$ and its fundamental frequency to $f_f$. In addition, $q_{i, min}$, $q_{i, max}$ and $\dot q_{i, min}$, $\dot q_{i, max}$ are the lower and upper bounds of the joint positions, and velocities, respectively. 

To handle the optimization, it is suggested to sample the trajectory with a fixed time step to averagely cover one Fourier series period, which makes the starting point and end point coincide with each other. The sampling frequency is set to $f_s = 20$\,Hz to reduce computational complexity, with the robotic controller operating at $100$\,Hz. The time duration of the trajectory execution is $T$, which, using (\ref{equ9b}), leads to $f_{s}T$ samples. Besides, \ref{equ9c}-\ref{equ9g} provide the constraints of the solution considering the robot's feasible configurations.

\subsection{Self-collision avoidance constraints} 
Self-collision during excitation trajectories execution must be addressed, especially considering the multitude of instruments that EEs in industrial and medical applications can hold. 
Sampling-based trajectory generation offers a simple solution.
However, enhanced performance with respect to condition number optimization is challenging. Sampling-based methods additionally might not find feasible solutions due to memory or time constraints; in such cases, collision-avoidance constraints represented by inequalities might be a more viable option.
One approach is to use the convex hull of the EE to model its original geometric shape. By using the convex hull, the complexity of collision detection calculations can be reduced. Note that the collision detection approach needs to be compatible with optimal excitation trajectory estimation, which implies that common approaches such as KD-trees cannot be used.

To further reduce computational complexity, two approaches can be considered. First, the main feature points on end effectors (MFPEE) are selected: a fast convex hull algorithm is applied and then the convex-hull points are refined by Gaussian Mixture Models (GMMs) which are utilized for point clustering, providing Gaussian distribution representations for each cluster. Within the Gaussian representation, the mean values can effectively encapsulate the distinctive MFPEE of each cluster:
\begin{equation}
    \mathbf{\mu_{K*}},  \mathbf{\sigma_{K*}} = \mathop{\arg\max}_{K} \sum_{i=1}^{N}\log(\sum_{k=1}^{K}\pi_{k}\mathfrak{N}(^{ee}\bar x_{M}|\mu_{k},  \sigma_{k})) 
    \label{equ9}
\end{equation}
where $\mathbf{\mu_{K*}}$ and $\mathbf{\sigma_{K*}}$ are the mean, and variance, parameters vectors of GMMs, respectively. Noticeably, $\mathbf{\mu_{K*}} = [\mu_{1},\mu_{2},...\mu_{K*}]$ and $\mathbf{\sigma_{K*}} = [\sigma_{1},\sigma_{2},...\sigma_{K*}]$ are defined with $\mu_{k}$ and $\sigma_{k}$ modelling the k-th Gaussian model in the GMMs. Parameter $\pi_k$ denotes the mixture weight of the k-th Gaussian model, which could be inferred from the Expectation-Maximization (EM) algorithm. The input points generated from the convex-hall-finding algorithm are denoted as $^{ee}\bar x_{M}$. From (\ref{equ9}), the original number of MFPEE, $N$, is reduced to $K*$ therefore improving on computational performance.

Second, we can approximate the links as ellipsoids and introduce them as virtual constraints. Then, a collision takes place if the MFPEE is within the constraints, represented as:
\begin{subequations}
\begin{equation}
    g_l = \mathbf{^{l}x_{e}^{T}}\;A\; \mathbf{^{l}x_{e}} -1
    \label{equ10a}
\end{equation}
\begin{equation}
    A\ = diag(\epsilon_{x}^{-2} \ \epsilon_{y}^{-2} \ \epsilon_{z}^{-2})
    \label{equ10b}
\end{equation}
\begin{equation}
    \mathbf{g_L} = [g_1, g_2, ..., g_L]^{T}<\mathbf{0_{L\times 1}}
    \label{equ10c}
\end{equation}
\end{subequations}
where constrains $g_l$ indicate the constraints built on the link $l$ of the robotic chain. Symbol $\mathbf{^{l}x_{e}}$ represents the point position of MFPEE with respect to the link $l$ frame. If there are $L$ links in the robotic chain, every link can be enumerated to check whether there are collisions. In practice, constraints of links that actually cannot contact EEs can be removed to improve computation efficiency. Parameters $\epsilon_{z}$,$\epsilon_{y}$,$\epsilon_{x}$ are the length of the major axis, intermediate axis, and minor axis of an ellipsoid, respectively. The size of the virtual ellipsoid can be set according to geometric measurements and estimates.
In order to account for uncertainties, the size of the ellipsoid should be larger than the original link shape in the potential collision area, and, based on experience, the medium parts of links in particular.

\section{Optimization Policy and Regression}
\label{Section4}
This part involves solving a regression for identification and applying a filter to enhance the identification quality. 
\subsection{Optimization Policy for Trajectories Generation}
The main challenge of the optimization is that it combines (\ref{equ9a}) - (\ref{equ9g}) and (\ref{equ10c}), which is not a classic SDP or QP problem as (\ref{equ10c}) is not linear. In fact, the problem we are called to solve is non-convex and periodic. Nonlinear optimization problems are usually very complex and variable, making it extremely difficult, and sometimes nearly impossible, to find a global optimal solution. Seeking a global optimal solution can involve a tremendous amount of computational cost and time. In many practical applications, finding a 'good' solution is often sufficient, rather than spending a great deal of time and resources to search for a possibly non-existent global optimal solution. 
Heuristics optimizers are less efficient and fail to scale with an increasing number of constraints. To address these issues, we consider multi-start optimization. A global reject-sampling-based sampler is utilized to generate initial points for the Interior Point OPTimizer (ipopt) optimizer. The initial points are taken randomly with the optimizer using Dual Gradient Descent (DGD) to avoid the calculations of Hessian matrix.

Empirically, (\ref{equ10c}) would not significantly change the solution with respect to the optimum given by (\ref{equ9a}) - (\ref{equ9g}). Therefore, in our implementation, a two-step optimization method is followed. For step 1, only (\ref{equ9a}) - (\ref{equ9g}) are considered. As a result, the $\mathop{nlpsol}$-based method can return a feasible solution after around 2000 iterations (for each initial start), while more than 15000 iterations required otherwise. The results of the $\mathop{nlpsol}$-based method are taken as an initial point for optimization considering (\ref{equ9a}) - (\ref{equ9g}) and (\ref{equ10c}).

After trajectory generation, all the $a_{i,l}, b_{i,l}$ are obtained. According to (\ref{equ9b}), target positions are given, while target velocities and accelerations can be derived from (\ref{equ9b}).

\subsection{Parameter regression}
To account for motion states' noise caused over recording, the measurements are filtered with a tracking-differentiator (TD) filter, which is mentioned in
\cite{ARDCHAN}.


%
We discretized our system using the following equations:
\begin{subequations}
\begin{equation}
    \hat q_1(t+1) = \hat q_1(t) + h\hat q_2(t)
    \label{equ12a}
\end{equation}
\begin{equation}
    \hat q_2(t+1) = \hat q_2(t) + h\;F_{h}(\hat q_1 (t)-\dot q_m (t), \hat q_2 (t))
    \label{equ12b}
\end{equation}
\end{subequations}
where where $\hat q_1(t)$ and $\hat q_2(t)$ are estimated velocities, and accelerations, respectively. Symbol $\dot q_{m} (t)$ denotes the measured velocities obtained from trajectory execution, with $\dot q_{raw} (t)$ being the tracked velocity signals.
$F_{h}$ indicates a piecewise function for detecting tracking signals (see details in \cite{ARDCHAN}). Variable $h$ represents the time step of discretization.

\subsection{Parameter Regression}
To regress the parameters, the problems can be represented using criteria and inequalities as follows:
\begin{subequations}
\begin{equation}
    \mathop{min}_{\theta} ||\tau_{raw}- \bar Y_{b}\theta_b|| 
    \label{equ13a}
\end{equation}
\begin{equation}
    s.t. \theta_{b,lb}(\mu) \leq \theta_b \leq \theta_{b,ub}(\mu) 
    \label{equ13b}
\end{equation}
\end{subequations}
where the $\theta_{lb}$ and $\theta_{ub}$ are the lower bound and upper bound of dynamic parameters $\theta$, respectively. The bounds can be estimated from the setting values of URDF of the LBR. Symbol $\mu$ denotes the nominal parameters containing mass $m$, inertia $I$, and the center of mass $c$, which could be obtained from URDF's properties. Function $\theta$ returns minimal inertia set values as a unique parameter to identify the states in a robotic dynamic system. The function is generated from symbol derivation (transferring (\ref{equ2}) to (\ref{equ3})) using CasADi.

\section{Experiments and Results}
\label{Section5}
We present experiments on the execution of the excitation trajectory, and on validation based on admittance control. 

\subsection{System Setup}

The experiments are conducted on the KUKA LBR Med7 R800, which is a serial robot with $n = 7$ revolute joints and $7\,$kg payload capability. A joint torque sensor and joint position encoder exist after each of its gearboxes. 
Attached to the EE is a fixture holding a rod with a diameter of $12$\,mm, mimicking an endoscope. 
We obtain the robot's URDF from the LBR-Stack \cite{huber2023lbrstack} and extract the links' geometric shapes, inertia parameters, and kinematic parameters. The LBR-Stack is further used to command the robot through ROS 2 topics at $100\,$Hz.


\begin{figure}[t]
	\centering
		\includegraphics[width=\columnwidth]{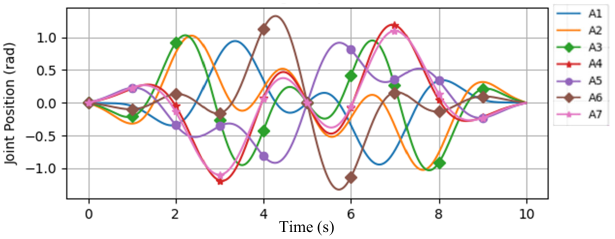}
	  \caption{Self-collision-free trajectory in joint space (Sec. \ref{sec:exciting_trajectory_identification_validation}).}
      \vspace{-10pt}
	  \label{figure2}
\end{figure}


\begin{figure}[t]
	\centering
		\includegraphics[width=0.95\columnwidth]{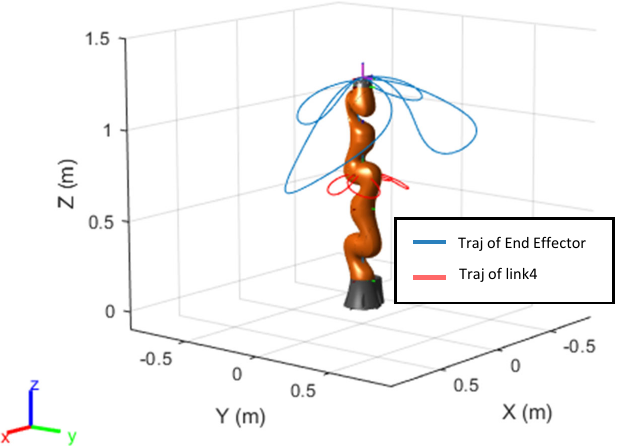}
	  \caption{Self-collision-free trajectory in Cartesian space (Sec. \ref{sec:exciting_trajectory_identification_validation}).}
   \vspace{-10pt}
	  \label{figure3}
\end{figure}

\begin{figure}[t]
	\centering
\includegraphics[width=\columnwidth]{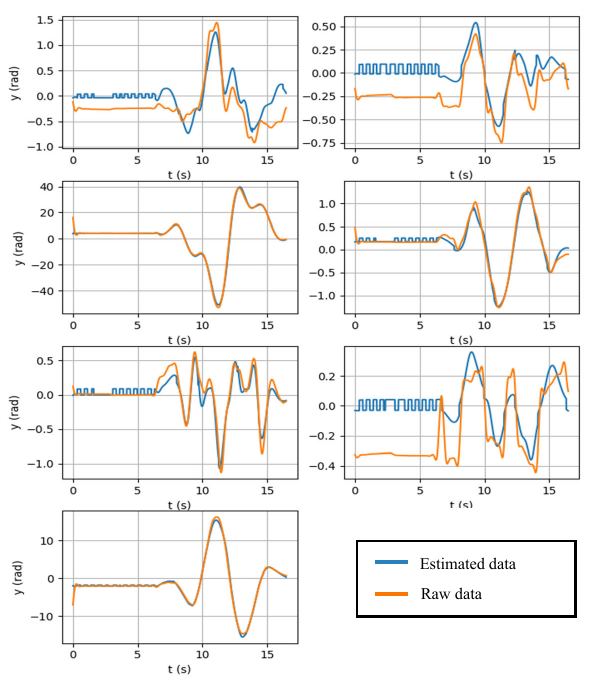}
	  \caption{The regression results of the excitation trajectory (Sec. \ref{sec:exciting_trajectory_identification_validation}).}
	  \label{figure4}
   \vspace{-10pt}
\end{figure}

\subsection{Exciting Trajectory Identification Validation}
\label{sec:exciting_trajectory_identification_validation}
Obtaining a feasible trajectory with the proposed method takes around 10 minutes, considering three random initial configurations. Using the condition number, however, requires more than 60 minutes, and also significantly higher memory usage.

The generated trajectory for estimating the robot's dynamic parameters, obtained by solving \ref{equ9a}-\ref{equ9g} and \ref{equ10a}-\ref{equ10c}, is shown in Fig.~\ref{figure2}. The optimization process leads to a collision-free trajectory. The conflict constraints in the NLP optimizer are below $10^{-3}$, which describes the absolute difference between the two sides of the inequalities.

The validation trajectory is then executed on the robotic system. Torques are measured over the entire trajectory and recorded as ground truth. The measured torques are compared to the predicted torques, which are obtained through the identified dynamic model. The maximum observed torque errors are $1.5\,$Nm for joint A1 and $0.5\,$Nm for joint A7, caused by limited motions in the two joints.

The condition number of $\mathbf{\bar Y}_b$ obtained via our excitation trajectory is 51, which is close to the condition number results without EEs (67 in \cite{sturz2017parameter} and 156.3 in \cite{xu2020dynamic}).
The condition number using heuristics (the memetic algorithm) in \cite{tika2020dynamic} to solve self-collision-based trajectory optimization) is 57890 within 40000 iterations, which is worse than our proposed method, both in time efficiency and value.

\begin{table*}[t]
\centering
\caption{NASA-TLX Table for the shared docking task, refer to Sec. \ref{sec:admittance_control}.}
\label{table1}
\begin{tabular}{|c|c|c|c|}
\hline
\multirow{2}{*}{\textbf{Dimension (weight)}} & \multicolumn{3}{c|}{\textbf{Score (0-100), lower is best}} \\ 
\cline{2-4} 
 & \textbf{Experimental Group} & \textbf{Control Group (CG)} & \textbf{CG with tool calibration} \\ \hline
Mental Demand (0.2) & $12.70\pm 5.85$&$63.80\pm 19.34$ & $21.80\pm 21.94$ \\ \hline
Physical Demand (0.2) & $12.50\pm 6.12$& $65.00\pm17.53$ & $22.70\pm 21.84$\\ \hline
Temporal Demand (0.1) & $19.90\pm 14.56$& $53.40\pm 27.63$ & $23.40\pm 20.32$\\ \hline
Performance (0.2) & $14.50\pm 9.53$&$63.70\pm 19.88$ & $23.60\pm 19.18$ \\ \hline
Effort (0.2) & $18.80\pm 9.39$& $58.90\pm 16.65$& $22.60\pm 19.11$\\ \hline
Frustration Level (0.1) & $13.50\pm 7.43$&$55.20\pm 24.06$& $21.70\pm 18.08$ \\ \hline
\textbf{Total NASA TLX Score} & $15.04\pm 8.38$ & $61.14\pm 19.85$& $22.65\pm 20.56$ \\ \hline
\end{tabular}
\vspace{-10pt}
\end{table*}

\subsection{Evaluation Considering Admittance Control}
\label{sec:admittance_control}
To assess the robustness of the regression-derived model, we apply an admittance controller to evaluate system drifting in the absence of external forces. To counter potential sensor noise, a dead zone threshold is established with the following parameters: $f_{th} = [2.0\, \text{N}, 2.0\, \text{N}, 2.0\, \text{N}, 0.1\, \text{Nm}, 0.1\, \text{Nm}, 0.1\, \text{Nm}]^{T}$.

\begin{figure}[t]
	\centering
	\includegraphics[width=0.95\columnwidth]{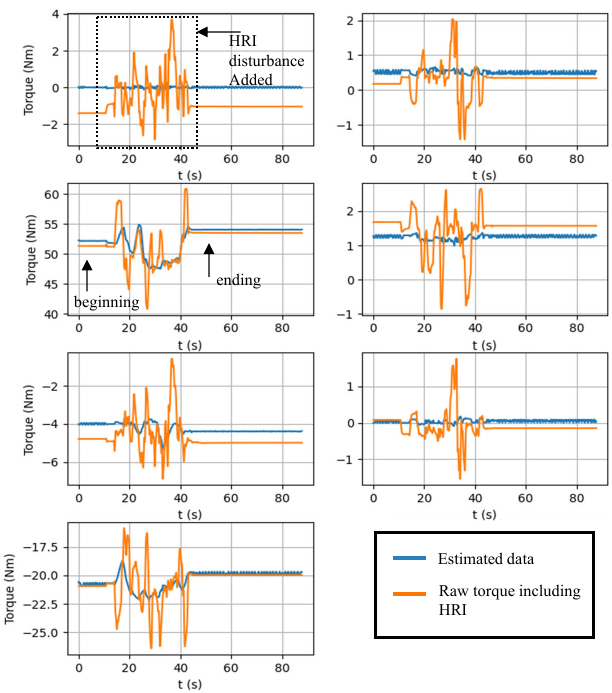}
	  \caption{Comparison between actual torques and estimated torques with disturbance by external pHRI forces, refer Sec. \ref{sec:admittance_control}.}
	  \label{figure5}
   \vspace{-10pt}
\end{figure}

The results of this evaluation are presented in Fig.~\ref{figure5}. During the experiment, a pHRI force acts upon the robot, which can be considered as an artificial disturbance in practical applications. In both, the initial and final stages of the experiment, this pHRI force is not present, as can be observed in the straight-line sections of the figure.

At the start and end of the experiment, biases caused by caused by static frictions can be observed.
Therefore, the beginning and ending periods — devoid of any external force disturbances — illustrate a force comparison when no payload is added. Utilizing this refined model allows the admittance controller to function with a reduced force dead zone threshold, making it less susceptible to drifting while facilitating quicker responses for transitioning external forces to co-manipulative movements.

\begin{figure}[t]
	\centering
		\includegraphics[scale=0.75]{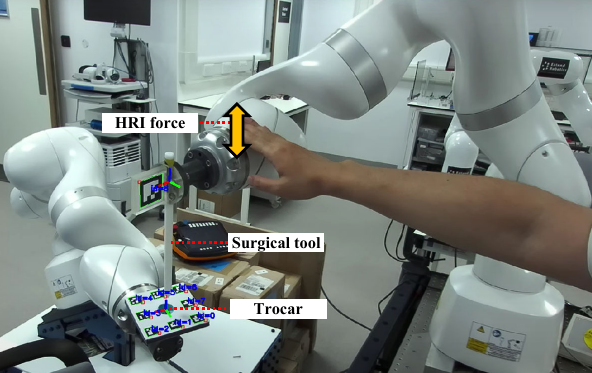}
	  \caption{Docking experiment set up. ArUco markers are used to identify the trocar's position with respect to the end-effector. The robot autonomously aligns the endoscope with the trocar. Users are then in charge of inserting and retracting the tool through admittance control. Refer to Sec. \ref{sec:admittance_control}.}
	  \label{figure6}
   \vspace{-10pt}
\end{figure}

Next, to practically evaluate the model's performance, we create a task drawn from the medical domain, i.e., co-manipulated docking of an endoscope through a trocar in laparoscopic surgery. Ten users participate in the docking test, shown in Fig.~\ref{figure6}. In a random fashion, they use the dynamic parameters identified by the proposed method (Column 1 in Table \ref{table1}), and the dynamic model obtained from KUKA, without tool calibration (Column 2 in Table \ref{table1}), and with tool calibration (Column 3 in Table \ref{table1}). Force thresholds and admittance matrix are kept the same, and no user has knowledge of which algorithm is used at what time. The task lets users control the insertion and retraction of the tool into the trocar, with lateral alignment controlled by visual servo. 

For visual servo control, we use a ZED 2i camera (Stereolabs, USA) to estimate real-time ArUco marker poses, both, on the trocar and the tool. The control policy is to reduce the distance between the two poses in task space.

NASA Task Load Index metrics \cite{law2020nasa} are gathered by all users to capture the mental, physical, temporal demand, etc. when either of the dynamic parameter sets are used. Lower scores signify better performance and acceptability. Weighted averaging of all indices results in a single metric (Total NASA TLX Score).

The results show our method offers a better estimation of dynamic parameters than the ones given through the FRI \cite{fri} from KUKA (both with tool calibration and without tool calibration) as it reduces the workload during co-manipulation. Although we observe an improvement in scores of NASA-TLX, both methods are good enough for pHRI and compatible with co-manipulation according to the user feedback. Critically though, our excitation trajectory takes significantly less time (10 s), which stands in comparison to 2 minutes for KUKA's tool calibration. Our method further provides full access to all inertias, whereas KUKA's tool calibration only yields the tool's center of mass and no other parameters are accessible to the user. Finally, KUKA's tool calibration is potentially harmful as no collision avoidance is guaranteed, whereas the proposed methods is safe to use at all times.

\section{Summary}
\label{Section6}

This paper presents  a novel excitation trajectory optimization for dynamic parameter estimation, incorporating self-collision and considering computational efficiency. An evaluation function based on the condition number upper bound is introduced, contributing to accurate and stable parameter estimation in system identification.
Admittance control is successfully applied in a medical robotics context, and the effect of the parameter estimation is evaluated in the context of a co-manipulation task using the NASA TLX questionnaire to quantify cognitive load on operators. Collectively, the introduced methodologies showed the potential to enhance safety and intuition in human-robot collaboration.






\bibliographystyle{IEEEtran}

\bibliography{IEEEabrv,BIB_xx-TIE-xxxx}\ 

\end{document}